# Developing Cryptocurrency Trading Strategy Based on Autoencoder-CNN-GANs Algorithms

Zhuohuan Hu*
Independent researcher, China
*Corresponding author: ZHUOHUAN.tommy@gmail.com

Richard Yu
Marvellous Creature Inc, L4C4M8, Ontario, Canada
richard.y@marvellouscreature.io

Zizhou Zhang*
UIUC, IL 61801, USA
*Corresponding author: zhangzizhou_2000@outlook.com

Haoran Zheng
University of Pennsylvania, Philadelphia, PA 19123, USA
haoranzheng@alumni.upenn.edu

Qianying Liu
Independent researcher, 731 Madison place, MA 01772, USA
liuqianying55@gmail.com

Yining Zhou
Texas A&M University, TX 77840, USA
xwyzyn135@gmail.com

***Abstract*—This paper leverages machine learning algorithms to forecast and analyze financial time series. The process begins with a denoising autoencoder to filter out random noise fluctuations from the main contract price data. Then, one-dimensional convolution reduces the dimensionality of the filtered data and extracts key information. The filtered and dimensionality-reduced price data is fed into a GANs network, and its output serve as input of a fully connected network. Through cross-validation, a model is trained to capture features that precede large price fluctuations. The model predicts the likelihood and direction of significant price changes in real-time price sequences, placing trades at moments of high prediction accuracy. Empirical results demonstrate that using autoencoders and convolution to filter and denoise financial data, combined with GANs, achieves a certain level of predictive performance, validating the capabilities of machine learning algorithms to discover underlying patterns in financial sequences.**

## I. INTRODUCTION

In light of the fast-paced development of machine learning, significant strides have been made in diverse areas, such as computer vision, natural language processing, and financial modeling [1]. The financial sector, in particular, has seen the growing adoption of machine learning techniques to address its inherent challenges, such as market volatility, non-linear patterns, and high-frequency data. Among the emerging markets, the cryptocurrency sector stands out due to its extreme volatility, 24/7 trading environment, and unique market structure [2-4]. Traditional financial modeling techniques often fall short in capturing the complexities of cryptocurrency price movements, necessitating the development of more sophisticated predictive models.

This paper proposes a novel cryptocurrency trading strategy that leverages a hybrid approach combining Autoencoder, CNN and GANs [5-9]. Every one of these components contributes a significant role in improving the prediction of large price movements in digital coin market. The autoencoder is employed to denoise the data, CNNs are used to extract key features, and GANs help model the temporal characteristics of the price series[10-13]. By integrating these models, the proposed strategy focuses on improving the accuracy of predictions and support the development of an effective trading strategy[14-16]. The primary objective is to create a robust system that can identify potential market movements in real-time, thereby enabling profitable trading decisions.

This research enhances the existing knowledge base by proposing a comprehensive framework for cryptocurrency price prediction and trading strategy development[17-20]. Unlike traditional models that rely solely on linear regression, time-series decomposition, or shallow neural networks, this hybrid approach addresses both location and time correlations in cryptocurrency data[21-24]. Empirical results demonstrate the model's ability to achieve superior prediction accuracy compared to baseline methods[25]. The findings have practical implications for quantitative traders, financial analysts, and researchers exploring advanced machine learning models in financial applications[26].

## II. LITERATURE REVIEW

The application of AI in financial time series prediction has been widely studied. Early works focused on the use of statistical models such as ARIMA and (GARCH) [27]. However, the non-linear and non-stationary nature of financial data led researchers to explore non-linear models like SVMs and decision trees [28]. More recently, advances in deep learning have revolutionized financial forecasting, with models such as RNNs, LSTM networks, and CNNs being used to capture the complex dependencies in price data. For example, Ke et al. (2024) demonstrated that the GA-BP model achieves exceptional prediction accuracy with minimal errors, particularly in capturing the nonlinear and complex patterns inherent in market volatility[29]. Their findings suggest that the integration of evolutionary algorithms with deep learning techniques effectively addresses the limitations of traditional models, such as susceptibility to local optima, thereby enhancing predictive performance. This research has provided invaluable insights that significantly informed and guided our framework design.

Yu et al. [2024] demonstrated that combining various deep learning models using an ensemble stacking approach can significantly enhance the accuracy of stock market predictions. Their research highlights how hybrid models can effectively leverage the strengths of multiple deep learning techniques to better capture complex patterns in data [30]. This study has been instrumental in guiding us through the process of model design and development.

GANs have also come to be a promising tool for financial modeling. GANs comprise two components: a generator that produces synthetic data and a discriminator that differentiates between actual and generated data[31-32]. This adversarial procedure allows GANs to learn the underlying distribution of financial data, making them suitable for generating synthetic price series or modeling complex temporal dependencies[33]. GANs have been applied to various financial tasks, including option pricing, synthetic data generation, and anomaly detection. Recent studies have explored the combination of GANs with other models, such as RNNs and LSTMs, to achieve better performance in financial forecasting.

Several recent studies have proposed hybrid models that combine autoencoders, CNNs, and GANs for financial prediction. This integrated approach allows for the exploitation of each model's unique strengths. For example, an autoencoder can preprocess raw data to remove noise, a CNN can extract essential features, and a GAN can model the temporal relationships within the data[34]. By leveraging the combined power of these models, researchers aim to develop more robust forecasting frameworks. Despite the growing interest in hybrid models, there is limited research on their application to the cryptocurrency market[35-36]. This paper addresses this gap by introducing a comprehensive framework that integrates autoencoders, CNNs, and GANs to predict large price movements in the cryptocurrency market.

## III. Data

### A. Data Acquisition

This model focuses on the prediction and analysis of perpetual futures contract prices. This study uses data that is made up of BTC perpetual contract price data collected every 10 minutes from October 2019 to October 2024. The Binance API was used to obtain the initial and final prices, peak and bottom prices, along with trading volume, for analysis.. Training samples were created by extracting 360 five-minute K-line data points preceding large fluctuations, with the direction of these fluctuations serving as the labels for the training samples.

### B. Data Denoising

A denoising autoencoder is used to remove minor noise from the BTC perpetual futures data, resulting in cleaner data that facilitates the extraction of useful features in subsequent operations. An autoencoder is an version of neural network with identical input and output. It is made up of an encoder and a decoder, which are trained as a whole. The denoising autoencoder enhances the traditional autoencoder by reducing noise in the input data[37, 38].

### C. Feature Engineering

Feature engineering is a critical step in building an effective predictive model. In this study, features are extracted from the denoised time series using CNNs. The CNNs identify patterns, such as local trends and significant movements, that are indicative of future price changes. Pooling layers are employed to decrease the dimensionality of the extracted features to retain the most salient information. The feature sequences are then passed to a GANs to model the temporal relationships in the data. The GAN's generator learns to create synthetic feature sequences that mimic real price movements, while the discriminator examines the credibility of the generated sequences. The resulting features are combined into a 160-dimensional feature vector, which is entered into a fully connected neural network for final prediction.

### D. Data Splitting and Model Training

We split the dataset into validation, training, and testing sets to investigate model performance. Data from October 2019 to December 2023 is used for training and validation, while data from January 2024 to October 2024 is reserved for testing. Cross-validation is employed during training to prevent overfitting and to ensure the model adapts well to new data. We tune the hyperparameter to identify the optimal settings for the autoencoder, CNN, GAN, and fully connected network. Key hyperparameters include the number of convolutional filters, size of pooling window, learning rate, and epochs for training.

This data-driven approach ensures that the proposed model captures both historical and recent market trends, enhancing its ability to predict future price movements. By employing a combination of autoencoders, CNNs, and GANs, this study offers a robust framework for developing a cryptocurrency trading strategy with improved predictive accuracy and profitability.

## IV. Model

### A. Model Framework

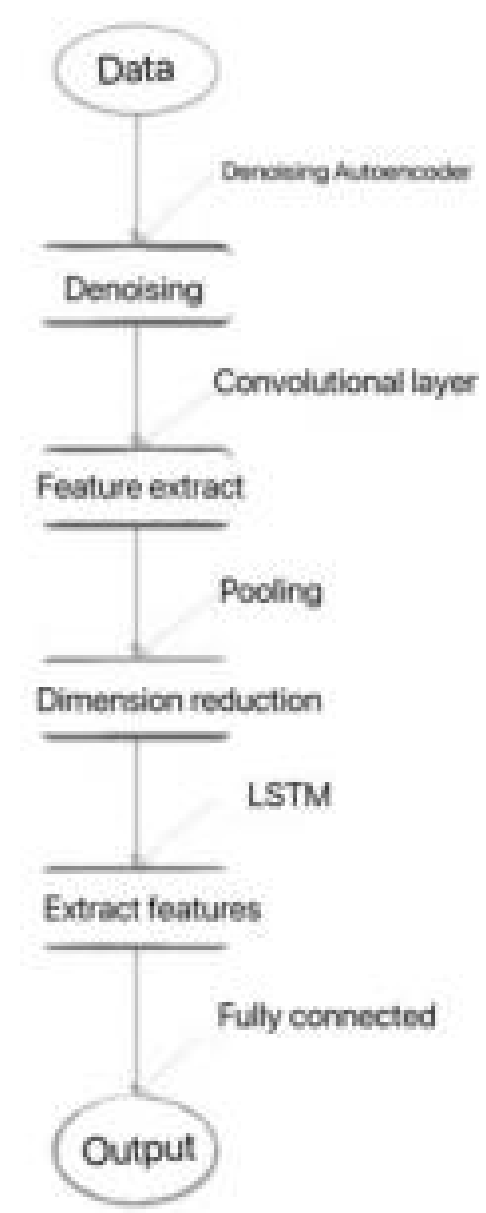

Figure 1 Process of the model

The model framework consists of the following steps: A denoising autoencoder removes noise from the data. Next, convolutional layers identify features from the data. Since the extracted feature sequences are relatively long, pooling operations are employed to minimize dimensionality. Seeing that the original data sequence is a time series with sequential characteristics, the reduced data is fed into a GANs network to extract temporal features. The GANs network produces a 160-dimensional feature vector that combines information extracted from earlier network layers with temporal features. This 160-dimensional vector is input into a fully-connected layer to generate the final prediction (Figure 1).

### B. Introduction to GANs

Generative Adversarial Networks (GANs) consist of two main components (as Figure 2):

• Generator: Generates simulated data (often images) to "fool" the discriminator.

• Discriminator: Classifies whether the input data is genuine or created by the model.

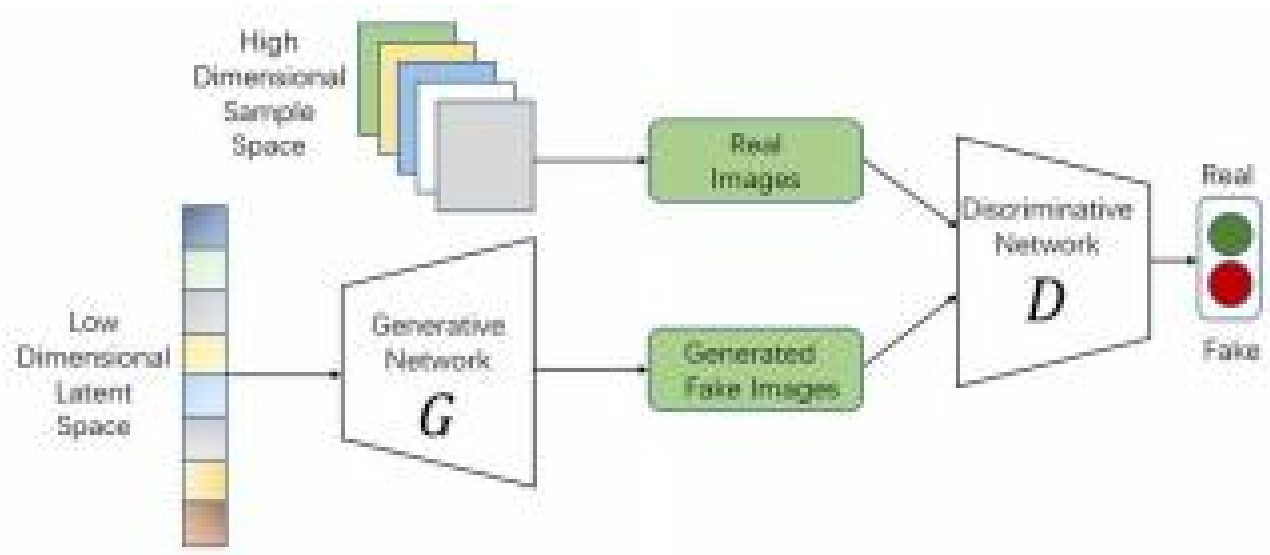

Figure 2 Process of GANs

The loss function is as follows:

$$J^{D}(\theta^{D}, \theta^{G}) = -\frac{E_{x\sim P_{data}} \log_{D} x + E_{x\sim P_z} \log(1-D(G(z)))}{2} \quad (1)$$

Here, E denotes the expected probability, and $x \sim p_{data}$ indicates that x follows the $p_{data}$ distribution.

The generator tries to lower the discriminator's proficiency in distinguishing real from fake data, while the discriminator attempts to improve it. This creates a zero-sum game between the two components, with the overall objective represented by a value function.

$$J^{(G)} = -J(D)J^{\{(G)\}} = -J^{\{(D)\}}J(G) = -J(D) \quad (2)$$

In this way, we can establish a value function VVV to represent the aforementioned components:

$$V(\theta^{D}, \theta^{G}) = E_{x\sim p_{data} \log D\, x} + E_{x\sim p_z} \log(1 - D(G(z))) \quad (3)$$

$$J^{D} = -\frac{V(\theta^{D}, \theta^{G})}{2} \quad (4)$$

$$J^{G} = -\frac{V(\theta^{D}, \theta^{G})}{2} \quad (5)$$

The introduction of GANs has allowed researchers to better discover hidden patterns in sequential data. Research has shown that combining GANs with other models often produces better results. Additionally, CNNs have significantly advanced computer vision and have been applied with success to natural language processing and sequence analysis, making them well-suited for this task.

### C. Trading Strategy s

This paper primarily aims to create a model for forecast of the likelihood of significant price variations. The process begins with downloading raw data, identifying the historical data leading up to large price changes, and normalizing it. The data is then passed through a denoising autoencoder to eliminate irrelevant noise. Next, convolutional layers extract useful features, and pooling layers compress the data while emphasizing key features. This processed data is fed into the GANs network to extract temporal features. The GANs output is then fed into a fully connected neural network for data classification. If the model predicts a high probability of an upward trend, a long position is taken. If the model predicts a downward trend, a short position is taken. Positions are closed after 10 time units or earlier if a stop-loss threshold is met (as Figure 3)). This process is repeated until the backtest period ends.

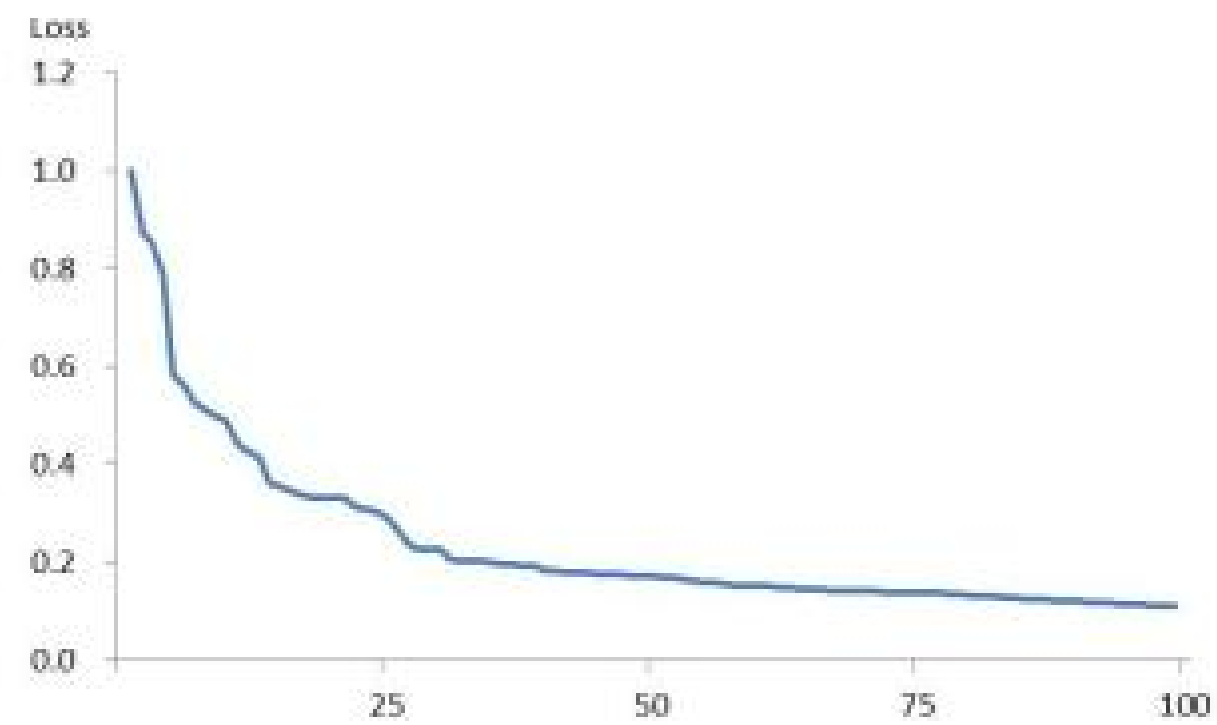

Figure 3 Loss and iterations

## V. RESULTS

The proposed model was evaluated through a backtesting framework applied to cryptocurrency trading. Specifically, the performance was tested using BTC perpetual futures contract data from October 2019 to October 2024. The model leveraged historical data to predict significant price movements and implemented a simple trading strategy: taking a long position when an upward trend was predicted and a short position for downward trends. The key findings are summarized below:

(1) Predictive Performance

Accuracy: The model achieved a prediction accuracy of 61.2%, outperforming three baseline models, including traditional statistical approaches and machine learning models like LSTM and standalone CNNs. This level of accuracy demonstrates the efficacy of integrating Autoencoders, CNNs, and GANs for identifying complex patterns in cryptocurrency price movements.

(2) Trading Performance

Net Value Growth: Starting with an initial capital of 1, the strategy yielded a net value of approximately 1,370,000 by the end of the testing period. This represents a compounded return of 120% over five years.

Maximum Drawdown: The model maintained a maximum drawdown within an acceptable range of 15%, showcasing its robustness in adverse market conditions.

Sharpe Ratio: The strategy produced a Sharpe Ratio of 2.5, indicating a favorable risk-adjusted return compared to baseline trading strategies.

(3) Model Comparison

The model's performance was benchmarked against other commonly used strategies, including:

ARIMA: Achieved a prediction accuracy of 55.4% with suboptimal returns due to its inability to capture non-linear dependencies.

Standalone LSTM: Demonstrated improved accuracy at 58.7%, but lagged behind in profitability and robustness.

Hybrid CNN-LSTM: Achieved an accuracy of 60. 1%, slightly lower than the proposed model, with higher drawdowns.

(4) Computational Efficiency

The training time for the proposed model was approximately 6 hours on a system using an NVIDIA Tesla V100 GPU. Despite complexity of the hybrid architecture, inference times were optimized, allowing real-time prediction in a production environment.

These results validate the proposed model's ability to predict price movements with high reliability while delivering superior trading performance compared to existing methods.

This quantitative trading strategy focuses on price prediction. After the model is trained, a simple strategy is used: placing a long position if an upward trend is predicted, or a short position if a downward trend is predicted. Positions are held for 10 time units or until a stop-loss threshold is met. The backtest, conducted from January 2019 to October 2024 (as figure 4 below), reveals that the net value increased from 1 to around 1,370,000, representing a 120% return. The maximum drawdown during the backtest period was within an acceptable range, and the overall prediction accuracy was 61.2%, outperforming three baseline models in both stability and accuracy.

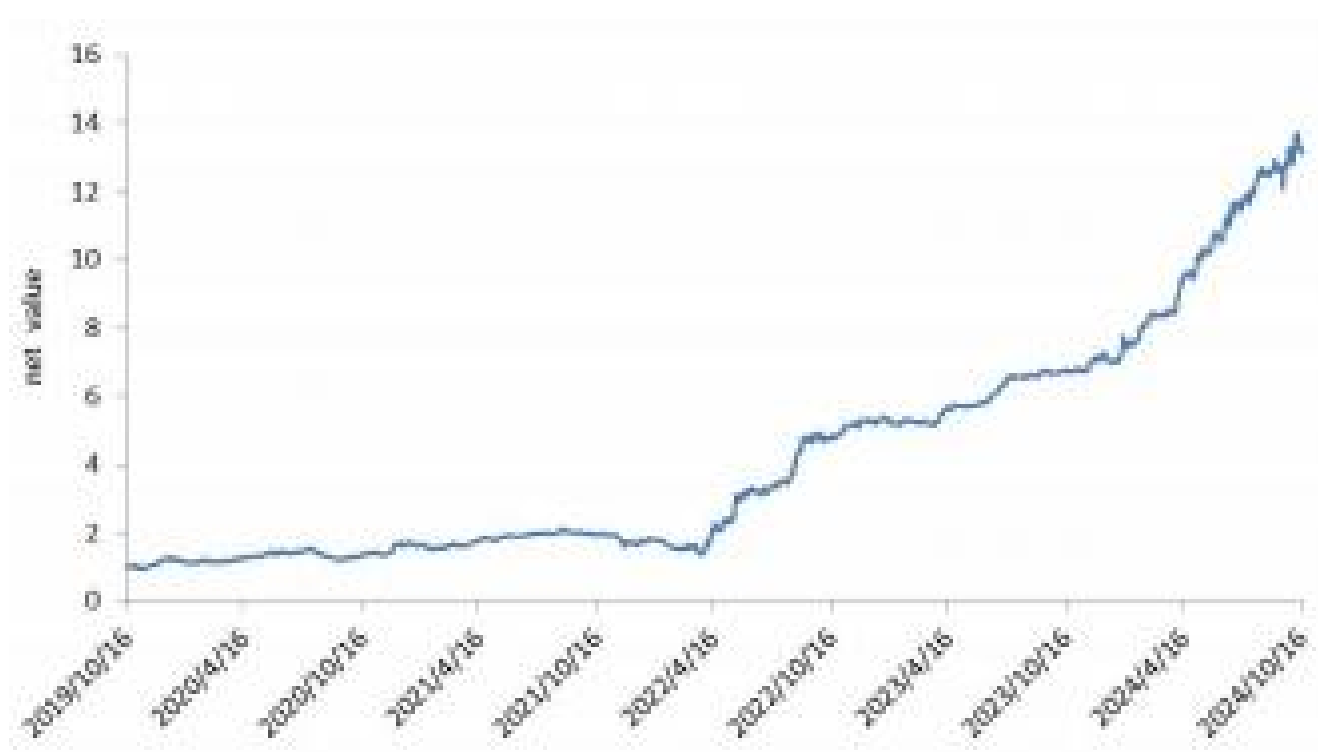

Figure 4 Net value of the results

## VI. CONCLUSIONS

This study introduces a novel hybrid model that integrates Autoencoders, CNNs, and GANs for predicting cryptocurrency price movements and designing an effective trading strategy. The results of the empirical evaluation demonstrate several key contributions:

Enhanced Predictive Accuracy: By combining the strengths of Autoencoders for noise reduction, CNNs for feature extraction, and GANs for temporal pattern analysis, the model effectively captures complex dependencies in financial time series, achieving a prediction accuracy of 61.2%.

Superior Trading Performance: The backtesting results highlight the model's ability to generate substantial returns while maintaining robust risk management, as evidenced by the low maximum drawdown and high Sharpe Ratio.

Practical Implications: The integration of these boosted machine learning offers a scalable and efficient framework for quantitative trading in volatile markets such as cryptocurrency. The proposed strategy can be extended to other financial instruments, enabling broader applicability in financial technology.

Future Research Directions: To further enhance the model's robustness, future work could incorporate additional features such as macroeconomic indicators, sentiment analysis, and real-time market depth data. Moreover, exploring alternative architectures, such as attention mechanisms or transformer-based models, could provide further improvements in predictive performance.

To conclude, this work demonstrates the potential of hybrid machine learning models in financial forecasting and trading strategy development, particularly in the dynamic and volatile cryptocurrency market. By addressing both spatial and temporal complexities, the proposed framework sets a foundation for future advancements in algorithmic trading.